\documentclass[10pt,twcolumn,journal]{IEEEtran}
\usepackage[T1]{fontenc}
\usepackage[latin9]{inputenc}
\usepackage{float}
\usepackage{amsmath}
\usepackage{amsthm}
\usepackage{lipsum}
\usepackage{amssymb}
\usepackage{graphicx}
\usepackage{setspace}
\usepackage[]{footmisc}
\usepackage[compact]{titlesec}
\usepackage[ruled]{algorithm2e}
\usepackage{algpseudocode}
\usepackage[unicode=true,
 bookmarks=true,bookmarksnumbered=true,bookmarksopen=true,bookmarksopenlevel=1,
 breaklinks=false,pdfborder={0 0 1},backref=false,colorlinks=false]
 {hyperref}

\hypersetup{pdftitle={Your Title},
 pdfauthor={Your Name},
 pdfborderstyle=,pdfpagelayout=OneColumn,pdfnewwindow=true,pdfstartview=XYZ,plainpages=false}

\makeatletter
\newcommand*{\rom}[1]{\expandafter\@slowromancap\romannumeral #1@}

\providecommand{\tabularnewline}{\\}
\floatstyle{ruled}
\newfloat{algorithm}{tbp}{loa}
\providecommand{\algorithmname}{Algorithm}
\floatname{algorithm}{\protect\algorithmname}
\renewcommand{\qedsymbol}{$\blacksquare$}

\theoremstyle{plain}
\newtheorem{thm}{\protect\theoremname}
\theoremstyle{remark}

\newtheorem{lemma}[thm]{Lemma}

\providecommand{\theoremname}{Theorem}
\usepackage{cite}
\author{Prabhu Babu and Petre Stoica
\thanks{Petre Stoica is with the Division of Systems and Control, Department of Information Technology, Uppsala University, Uppsala, Sweden 75237 and Prabhu Babu is with the Centre for Applied Research in Electronics, Indian Institute of Technology, Delhi 110016, India (email: ps@it.uu.se, Prabhu.Babu@care.iitd.ac.in).
Petre Stoica's work was supported in part by the Swedish Research Council (VR grants 2017-04610,  2016-06079, and 2021-05022).}
}
\makeatother

\providecommand{\remarkname}{Remark}
\providecommand{\theoremname}{Theorem}

\begin{document}
\title{Fair principal component analysis (PCA): minorization-maximization algorithms for Fair PCA, Fair Robust PCA and Fair Sparse PCA}
\maketitle
\begin{abstract}
In this paper we propose a new iterative algorithm to solve the fair PCA (FPCA) problem. We start with the max-min fair PCA formulation originally proposed in \cite{samadi1} and derive a simple and efficient iterative algorithm which is based on the minorization-maximization (MM) approach. The proposed algorithm relies on the relaxation of a semi-orthogonality constraint  which is proved to be tight at every iteration of the algorithm. The vanilla version of the proposed algorithm requires solving a semi-definite program (SDP) at every iteration, which can be further simplified to a quadratic program by formulating the dual of the surrogate maximization problem. We also propose two important reformulations of the fair PCA problem: a) fair robust PCA - which can handle outliers in the data, and b) fair sparse PCA - which can enforce sparsity on the estimated fair principal components.
The proposed algorithms are computationally efficient and monotonically increase their respective design objectives at every iteration. An added feature of the proposed algorithms is that they do not require the selection of any hyperparameter (except for the fair sparse PCA case where a penalty parameter that controls the sparsity has to be chosen by the user). We numerically compare the performance of the proposed methods with two of the state-of-the-art approaches on synthetic data sets and a real-life data set.
\end{abstract}

\section{Introduction and problem formulations}
Principal component analysis (PCA) is one of the most widely used unsupervised dimensionality reduction technique \cite{PCA1}, which has found applications, among others, in image processing \cite{PCAimage1, PCAimage2}, signal processing \cite{PCASP1, PCASP2}, finance \cite{PCAfinance}, and biology \cite{PCAbio1,PCAbio2}. Given the data, PCA learns the underlying low dimensional subspace by minimizing the reconstruction error. Let $\mathbf{Y} = [\mathbf{y}_1,\mathbf{y}_2, \cdots, \mathbf{y}_N]\in \mathbb{R}^{n\times N}$ denote the data samples, where $n$ and $N$ ($n\leq N$) represent the dimension and the number of data samples, respectively. Without loss of generality, we assume that the data samples are mean centered, i.e. $\sum_{i=1}^N \mathbf{y}_i = \boldsymbol{0}$. Mathematically, PCA solves the following problem to find the low rank matrix $\hat{\mathbf{Y}}$:
\begin{equation}
\begin{array}{ll}
\mathbf{\hat{Y}}=\textrm{arg}\,\underset{\mathbf{Z}}{\min} \: \displaystyle \lVert\mathbf{Y}-{\mathbf{Z}}\rVert \;\;  \textrm{s.t.}  \;\; \textrm{rank}({\mathbf{Z}}) =r\leq n
\end{array}
\label{Rank}
\end{equation}
where $\lVert\cdot\rVert$ denotes the Frobenius norm and $r$ denotes the rank of the low-dimensional subspace, which is usually pre-specified. 
Although the problem in (\ref{Rank}) is non-convex (due to the presence of the rank constraint), its solution can be computed analytically using the singular value decomposition of $\mathbf{Y}$ :\\
\begin{equation*}
\mathbf{Y}=\begin{bmatrix} \underbrace{\mathbf{U}}_{r} & \underbrace{\mathbf{\bar{U}}}_{n-r} \end{bmatrix} \begin{bmatrix} \boldsymbol{\Sigma} & 0 \\ 0 & \bar {\boldsymbol{\Sigma}}  \end{bmatrix} 
\begin{bmatrix} 
\begin{aligned}
     \left. \mathbf{V}^T \right\} & \; {r} \\
     \left. \bar{\mathbf{V}}^T \right\} & \; {n-r}\\
\end{aligned}
\end{bmatrix} \\
\end{equation*}
where the diagonal matrix $\begin{bmatrix} \boldsymbol{\Sigma} & 0 \\ 0 &\bar {\boldsymbol{\Sigma}}  \end{bmatrix}$ contains the singular values in decreasing order and the unitary matrices
$\begin{bmatrix} {\mathbf{U}}\; {\mathbf{\bar{U}}} \end{bmatrix} $ and $\begin{bmatrix}  \begin{aligned}{\mathbf{V}^{T}} \end{aligned}  \\   \begin{aligned}{\bar{\mathbf{V}}^{T}} \end{aligned}  \end{bmatrix}$ comprise the corresponding left and right singular vectors. Then, the low rank approximation $\hat{\mathbf{Y}}$ is given by
\begin{equation*}
\hat{\mathbf{Y}}=\mathbf{U}\boldsymbol{\Sigma}\mathbf{V}^{T}=\mathbf{U}\mathbf{U}^{T}\mathbf{Y}
\label{rank_app}
\end{equation*}
The reconstruction error is given by
\begin{equation}
\begin{array}{ll}
\displaystyle {\lVert\mathbf{Y}-\hat{\mathbf{Y}}\rVert^{2}}={\lVert(\mathbf{I}-\mathbf{U}\mathbf{U}^{T})\mathbf{Y}\rVert^{2}}&=\textrm{Tr}[\mathbf{Y}^{T}(\mathbf{I}-\mathbf{U}\mathbf{U}^{T})\mathbf{Y}] \\
&= \textrm{const.}-\textrm{Tr}(\mathbf{U}^{T}\mathbf{R}\mathbf{U}) \label{eq:mod}
\end{array}
\end{equation}
where $\mathbf{R}\triangleq\displaystyle \mathbf{Y}\mathbf{Y}^{T}$ is the sample covariance matrix.\\ 
Using (\ref{eq:mod}), the problem in (\ref{Rank}) can be reformulated as:
\begin{equation}
\begin{array}{ll}
{\max\limits_\mathbf{U}}  &  \textrm{Tr}({\mathbf{U}^{T}}\mathbf{R}\mathbf{U})\;\; \textrm{s.t.} \;\;\mathbf{U}^{T} \mathbf{U}=\mathbf{I}
\end{array} \label{eq:prob}
\end{equation}
The maximizer of (\ref{eq:prob}) is given by the $r$ principal eigenvectors of $\mathbf{R}$, and the cost function in (\ref{eq:prob}) is the Rayleigh quotient which can also be interpreted as the projection variance. However, in real-life applications the data samples come from diverse classes, so maximizing the average variance as in (\ref{eq:prob}) cannot ensure fairness among the classes. In other words, the matrix $\mathbf{U}$ determined by solving (\ref{eq:prob}) may be a good fit for some data samples and a poor fit for others. To mitigate this problem, a worst case approach named Fair PCA (FPCA) was recently proposed, which solves the following problem. Let us assume that the data samples are from $K$ different classes with the matrix $\mathbf{Y}_k$ containing the samples belonging to class $k$. The corresponding sample covariance matrix is $\mathbf{R}_{k} \triangleq \mathbf{Y}_k \mathbf{Y}_k^T$. Then the FPCA problem can be stated as \cite{samadi1}: 
\begin{equation}
\begin{array}{ll}
{\max\limits_\mathbf{U}}   &   \min\limits_{k\in [1,K]} \;\;\textrm{Tr}(\mathbf{U}^{T}\mathbf{R}_{k}\mathbf{U})\;\; \textrm{s.t.}\;\; \mathbf{U}^{T}\mathbf{U}=\mathbf{I}\\
\end{array} \label{eq:fpca}
\end{equation}
The solution to (\ref{eq:fpca}) will give the best fit for the least favourable class, and in general will balance the fitting with potentially many classes having identical variances $\{\textrm{Tr}(\mathbf{U}^{T}\mathbf{R}_{k}\mathbf{U})\}$. The FPCA problem in (\ref{eq:fpca}) is hard to solve due to the presence of the inner minimization operator and the constraint. In \cite{samadi1} and \cite{samadi2}, the authors have proposed algorithms for solving (\ref{eq:fpca}) using a semidefinite relaxation (SDR) technique. However the relaxation employed by SDR may not be tight for all problem dimensions and therefore may yield suboptimal solutions (except in the case $K=2$ where the relaxation is always tight as proved in \cite{samadi1}). In \cite{pareto}, the authors have proposed an alternative formulation of FPCA as a multi-objective optimization problem and suggested an iterative method to arrive at solutions spanning the Pareto frontier. Finally \cite{ami} has proposed a sub-gradient based algorithm to solve the FPCA problem. However, as for any gradient approach, the algorithm of \cite{ami} requires a careful choice of the step-size: failing to choose a good step-size may result in either convergence to a suboptimal solution or a very slow convergence.

In this paper, we propose an iterative algorithm to solve the FPCA problem, which is based on the minorization-maximization (MM) approach. Unlike some of the aforementioned algorithms, the proposed algorithm monotonically increase the cost function (i.e. $\min\limits_{k}\{\textrm{Tr}(\mathbf{U}^{T}\mathbf{R}_{k}\mathbf{U})\}$) and it is guaranteed to converge to a stationary point of the maximization problem in (\ref{eq:fpca}). An added feature of our algorithm is that it does not require the selection of any tuning parameter. 

In the presence of outliers, the solution to the problem formulated as in \eqref{eq:fpca} can be significantly affected. To tackle the outliers, we propose the following robust version of the fair PCA problem (hereafter called FRPCA):
\begin{equation}
\begin{array}{ll}
{\max\limits_\mathbf{U}} & \min\limits_{k\in [1,K]} \;\;\ \|\mathbf{U}^{T}\mathbf{Y}_{k}\|_{1}\;\; \textrm{s.t.}\;\; \mathbf{U}^{T}\mathbf{U}=\mathbf{I}\\
\end{array} \label{eq:fpca_robust}
\end{equation}
The $\ell_{1}$ norm based fitting criterion in (\ref{eq:fpca_robust}) is inspired by the L$1$  PCA formulation proposed in \cite{RPCA} which is the robust counterpart of the vanilla PCA and has been found to be quite effective in handling outliers. The proposed MM algorithm for (\ref{eq:fpca}) can be relatively easily adapted to solve the FRPCA problem in \eqref{eq:fpca_2}.

Finally, we also consider an extension of the proposed MM based algorithm for (\ref{eq:fpca}) to cases in which the sparsity of the fair PCA solution is important. To do this, we consider the following penalized problem:
\begin{equation}
\begin{array}{ll}
{\max\limits_\mathbf{U}}   &   \min\limits_{k\in [1,K]} \;\;\ \textrm{Tr}(\mathbf{U}^{T}\mathbf{R}_{k}\mathbf{U})-\lambda \|\mathbf{U}\|_{1}\;\; \textrm{s.t.}\;\; \mathbf{U}^{T}\mathbf{U}=\mathbf{I}\\
\end{array} \label{eq:fpca_2}
\end{equation}
where $\|\mathbf{U}\|_{1}$ denotes the sum of the absolute values of the elements of $\mathbf{U}$ and $\lambda$ is a prespecified penalty that controls the sparsity of $\mathbf{U}$. The formulation in (\ref{eq:fpca_2}) is motivated by the literature on sparse PCA \cite{sparsePCA1,sparsePCA2,sparsePCA3} where the $\ell_{1}$ norm of $\mathbf{U}$ is usually added as a penalty to the objective in (\ref{eq:prob}) to promote sparsity and hence help identify the significant contributing elements of the estimated principal components. As we will show later the proposed method for (\ref{eq:fpca}) can be extended to deal with the problem (\ref{eq:fpca_2}) in a straightforward manner. \\
The main contributions of this paper can be summarized as follows:\\
1) We propose a hyperparameter free algorithm based on the MM approach for solving the FPCA problem. The proposed algorithm has a monotonic convergence behavior and can be efficiently implemented.\\
2) We present a modification of the proposed algorithm that can deal with outliers in the data. More precisely, we formulate the fair robust PCA (FRPCA) problem and show how the MM algorithm for the FPCA approach can be modified to solve the FRPCA problem.\\
3) We also present an extension of the proposed algorithm (abbreviated as FSPCA) that can handle a sparsity promoting penalty in the FPCA problem.\\
4) Finally, we present several numerical results on both synthetic data sets and a real-life data set and compare the performance of the proposed algorithm with that of two state-of-the-art approaches.

\textit{Notations}: We use bold uppercase (e.g., $\mathbf{Y}$), bold lower case (e.g., $\mathbf{x}$) for matrices and vectors, respectively and italic letters (e.g., $x$) for scalars. The superscript $t$ of $\mathbf{x}^t$ denotes the iteration index. For any symmetric matrices $\mathbf{A}$ and $\mathbf{B}$, $\mathbf{A} \succcurlyeq \mathbf{B}$ means that $\mathbf{A}-\mathbf{B}$ is positive semidefinite. The variables $\mathbf{I}_{r}$ and $\mathbf{I}_{n}$ denote the identity matrices of dimension $r \times r$ and $n \times n$. $\mathbf{A}^{\frac{1}{2}}$ denotes the Hermitian square-root of the positive definite matrix $\mathbf{A}$ and sgn($x$) denotes the sign of the scalar $x$. Finally, $\|\mathbf{U}\|_1$ denotes the sum of the absolute values of the elements of the matrix $\mathbf{U}$.

A brief outline of the paper is as follows. In Section \rom{2} we first discuss the general MM approach for a maximization problem, and then the MM approach for max-min problems. In Section \rom{3} we present the proposed algorithm for the FPCA problem and discuss an efficient way to implement it. In Section \rom{4} we consider an extension of the proposed method to deal with fair robust PCA and fair sparse PCA problems. In Section \rom{5}, we present several numerical simulation results on both synthetic and real-life data and compare the proposed method with two state-of-the-art methods. Finally, we conclude the paper in Section \rom{6}.

\section{MM Primer}
In this section, we first present the main steps of the MM approach for maximization problems. Later we discuss how the MM approach can be extended to deal with max-min problems, which are of interest in this paper.
\subsection{MM for max problems}
Consider the following constrained maximization problem:
\begin{equation}\label{chapter2gen}
\begin{array}{ll}
\underset{\mathbf{x} \in \chi}{\rm max} \: f(\mathbf{x})
\end{array}
\end{equation}
where $\mathbf{x}$ is the optimization variable, $f(\mathbf{x})$ denotes the objective function and $\chi$ is a constraint set. An MM-based algorithm solves the above problem by first constructing a surrogate function $g(\mathbf{x}|\mathbf{x}^{t})$ which strictly lower bounds the objective function $f(\mathbf{x})$ at the current iterate $\mathbf{x}^{t}$. Then in the second step, the surrogate function is maximized to get the next iterate i.e., 
\begin{equation}  \label{chapter2eq:mmc}
\mathbf{x}^{t+1} \in \underset{\mathbf{x} \in \chi}{\rm arg\:max} \: g\left(\mathbf{x}|\mathbf{x}^{t}\right)
\end{equation}
The above two steps are repeated until the algorithm converges to a stationary point of the problem in (\ref{chapter2gen}).
The function $g\left(\mathbf{x}|\mathbf{x}^{t}\right)$ qualifies to be a surrogate function if it satisfies the following conditions: 
\begin{equation}\label{chapter2eq:mmb}
g\left(\mathbf{x}|\mathbf{x}^{t}\right) \leq f\left(\mathbf{x}\right)\,\forall\, \mathbf{x} \in \chi 
\end{equation}
\begin{equation}  \label{chapter2eq:mma}
g\left(\mathbf{x}^{t}|\mathbf{x}^{t}\right) = f\left(\mathbf{x}^{t}\right) 
\end{equation}
To summarize, the major steps of the MM approach are:
\begin{enumerate}
    \item Start with a feasible $\mathbf{x}^{0}$ and set $t=0$.
     \item Construct a minorizing function $g\left(\mathbf{x}|\mathbf{x}^{t}\right)$ of $f(\mathbf{x})$ at $\mathbf{x}^{t}$
     \item Obtain $\mathbf{x}^{t+1} \in \underset{\mathbf{x} \in \chi}{\rm arg\:max} \: g\left(\mathbf{x}|\mathbf{x}^{t}\right)$.
     \item If $\frac{|f(\mathbf{x}^{t})-f(\mathbf{x}^{t+1})|}{|f(\mathbf{x}^{t})|} < \epsilon$, where $\epsilon$ is some prespecified convergence threshold, exit; otherwise set $t=t+1$ and go to Step 2.
\end{enumerate}
It is straightforward to show that the MM steps monotonically increase the objective at every iteration i.e.
\begin{equation*}
    f(\mathbf{x}^{t+1}) \geq g\left(\mathbf{x}^{t+1}|\mathbf{x}^{t}\right) \geq g\left(\mathbf{x}^{t}|\mathbf{x}^{t}\right)=f(\mathbf{x}^t)  
\end{equation*}
The first inequality and the third equality follow from (\ref{chapter2eq:mmb}) and (\ref{chapter2eq:mma}) and second inequality from 
(\ref{chapter2eq:mmc}).
\subsection{MM for max-min problems}
Consider the following max-min optimization problem: 
\begin{equation} \label{chapter2minp}
\begin{array}{ll}
\underset{\mathbf{x} \in \mathcal{X} }{\rm max} \: \left\{f(\mathbf{x}) \triangleq \underset{i =1,2, \cdots, K}{\rm min}\:{f}_{i}(\mathbf{x})\right\}
\end{array}
\end{equation}
A possible surrogate function $g(\mathbf{x}|\mathbf{x}^{t})$ for the max-min problem is as follows: 
\begin{equation}\label{chapter2surrogateminmax}
\begin{array}{ll}
g(\mathbf{x}|\mathbf{x}^{t})= \underset{i =1,2, \cdots, K}{\rm min}\: {g}_{i}(\mathbf{x}|\mathbf{x}^{t}) 
\end{array}
\end{equation}
where each ${g}_{i}(\mathbf{x}|\mathbf{x}^t)$ is a tight lower bound on ${f}_{i}(\mathbf{x})$ at $\mathbf{x}^t$. The individual surrogates ${g}_{i}(\mathbf{x})$ satisfy the following conditions: 
\begin{equation}\label{chapter2cond1}
\begin{array}{ll}
{g}_{i}(\mathbf{x}^{t}|\mathbf{x}^{t}) ={f}_{i}(\mathbf{x}^{t})
\end{array}
\end{equation}
\begin{equation}\label{chapter2cond2}
\begin{array}{ll}
{g}_{i}(\mathbf{x}|\mathbf{x}^{t}) \leq {f}_{i}(\mathbf{x})
\end{array}
\end{equation}
One can easily show that the surrogate function ${g}(\mathbf{x}|\mathbf{x}^{t})$  defined in (\ref{chapter2surrogateminmax}) satisfies the conditions (\ref{chapter2eq:mmb}) and (\ref{chapter2eq:mma}):
\begin{equation}\label{chapter2mme}
\begin{array}{ll}
{g}_{i}(\mathbf{x}|\mathbf{x}^{t}) \leq {f}_{i}(\mathbf{x})  \implies &\underset{i =1,2, \cdots, K}{\rm min}\:{g}_{i}(\mathbf{x}|\mathbf{x}^{t})\\ &\leq \underset{i =1,2, \cdots, K}{\rm min}\:{f}_{i}(\mathbf{x})\\
&\implies g(\mathbf{x}|\mathbf{x}^{t}) \leq f(\mathbf{x}). 
\end{array}
\end{equation}
and 
\begin{equation}\label{chapter2mmd}
\begin{array}{ll}
g(\mathbf{x}^{t}|\mathbf{x}^{t}) = \underset{i =1,2, \cdots, K}{\rm min}\: {g}_{i}(\mathbf{x}^t|\mathbf{x}^{t})  = \underset{i =1,2, \cdots, K}{\rm min}\: {f}_{i}(\mathbf{x}^{t})  = f(\mathbf{x}^{t})
\end{array}
\end{equation}
Similar to the general MM case, here also one can show that the iterates $\left\{\mathbf{x}^{t}\right\}$ increase the objective function $f(\mathbf{x})$ monotonically and converge to a stationary point. We refer the reader to \cite{mohamad} \cite{pulmar} for application of MM to max-min problems in communications and radar, and to \cite{MM} for a detailed discussion of the MM approach including different ways of finding the surrogate function in the context of different applications.
\section{FPCA}
In this section, we will derive the MM algorithm for the problem in (3). For the sake of convenience, we restate the FPCA problem:
\begin{equation}
\begin{array}{ll}
 {\max\limits_\mathbf{U}}   &   \min\limits_{k\in [1,K]} \;\;f_{k}(\mathbf{U})\;\; \textrm{s.t.}\;\; \mathbf{U}^{T}\mathbf{U}=\mathbf{I}\\
\end{array} \label{P1}
\end{equation}
where $f_{k}(\mathbf{U})\triangleq\textrm{Tr}(\mathbf{U}^{T}\mathbf{R}_{k}\mathbf{U})$. Before we venture into solving (\ref{P1}), we state and prove a lemma which will help us in the development of the proposed algorithm.
\begin{lemma}\label{lemma 1}
The non-convex semi-orthogonality constraint $\mathbf{U}^{T}\mathbf{U}=\mathbf{I}$ in (\ref{P1}) can be relaxed to $\mathbf{U}^{T}\mathbf{U}\preccurlyeq \mathbf{I}$ and the global maximizer of the relaxed problem will satisfy the constraint in (\ref{P1}).
\end{lemma}
\IEEEproof{Observe that in (\ref{P1}) $(\text{for any $k$})$
\begin{equation*}
\begin{array}{ll}
    \textrm{Tr}(\mathbf{U}^{T}\mathbf{R}_{k}\mathbf{U})=\textrm{Tr}(\mathbf{R}_{k}\mathbf{U}\mathbf{U}^{T})&=\textrm{Tr}(\mathbf{R}_{k}\mathbf{U}\mathbf{QQ}^{T}\mathbf{U}^{T}) \\
    & \hspace{-20pt}\textrm{ for any $\mathbf{Q}$ such that $\mathbf{QQ}^{T}=\mathbf{I}$}
    \end{array}
\end{equation*}

Thus for any $\mathbf{U}$ satisfying the relaxed constraint we can choose $\mathbf{Q}$ so that $\mathbf{Q}^T \mathbf{U}^{T}\mathbf{U}\mathbf{Q}=\text{diagonal}\triangleq\boldsymbol{\Lambda} \preccurlyeq \mathbf{I}$. Because $\mathbf{U}^{T}\mathbf{U}$ and $\mathbf{UU}^{T}$ have same non-zero eigenvalues, we have that:
\begin{equation}
\mathbf{UU}^{T}=\mathbf{V}\boldsymbol{\Lambda}\mathbf{V}^{T}
    \label{Eq2}
\end{equation}
where $\mathbf{V}$ contains the principal eigenvectors of $\mathbf{UU}^{T}$ and $\mathbf{V}^{T}\mathbf{V}=\mathbf{I}$. Using (\ref{Eq2}) we have :
\begin{equation*}
\begin{array}{ll}
    \textrm{Tr}(\mathbf{U}^{T}\mathbf{R}_{k}\mathbf{U})=\textrm{Tr}((\mathbf{V}^{T}\mathbf{R}_{k}\mathbf{V})\boldsymbol{\Lambda})&=\sum\limits_{i=1}^{r}(\mathbf{V}^{T}\mathbf{R}_{k}\mathbf{V})_{ii}\boldsymbol{\Lambda}_{ii} \\
    &\leq \sum\limits_{i=1}^{r}(\mathbf{V}^{T}\mathbf{R}_{k}\mathbf{V})_{ii}
\end{array}
\end{equation*}
 Therefore all functions in (\ref{P1}) are larger for $\boldsymbol{\Lambda}=\mathbf{I}$ than for $\boldsymbol{\Lambda} < \mathbf{I}$, and this implies that the global maximizer of (\ref{P1}) under the constraint $\mathbf{U^{T}U}\preccurlyeq \mathbf{I} $ will satisfy the constraint in (\ref{P1}). Therefore the relaxation $\mathbf{U^{T}U}\preccurlyeq \mathbf{I}$ does not effect the solution of (\ref{P1}).
 \hfill \qedsymbol}\\

Using Lemma\,\ref{lemma 1} we replace the problem in (\ref{P1}) with the following relaxed problem:
\begin{equation}
\begin{array}{ll}
{\max\limits_\mathbf{U}}   &   \min\limits_{k\in [1,K]} \;\;\textrm{Tr}(\mathbf{U}^{T}\mathbf{R}_{k}\mathbf{U})\;\; \textrm{s.t.}\;\; \mathbf{U}^{T}\mathbf{U} \preccurlyeq \mathbf{I}\\
\end{array} \label{P2}
\end{equation}
The constraint in (\ref{P2}) is convex as it can be reformulated as a linear matrix inequality (\cite{boyd}):
\begin{equation*}
    \begin{bmatrix}
         \mathbf{I}_{r} & \mathbf{U}^{T} \\ \mathbf{U} & \mathbf{I}_{n}  
    \end{bmatrix}\succcurlyeq 0
\end{equation*}
However, the maximization problem in (\ref{P2}) is non-convex as the objective (for any $k$) is a convex quadratic function in $\mathbf{U}$ and the presence of $\text{min}$ operator further adds to complications. We resort to the MM approach to solve (\ref{P2}). Following the MM steps listed in Section II.B, each convex quadratic function in (\ref{P2}) can be lower bounded via its tangent hyperplane at $\mathbf{U}^{t}$. Doing so gives us an MM surrogate for the objective in (\ref{P2}). For given $\mathbf{U}^{t}$ and for any $k$, we have
\begin{equation*}
\begin{array}{ll}
f_{k}(\mathbf{U})&=\textrm{Tr}(\mathbf{U}^{T}\mathbf{R}_{k}\mathbf{U}) \geq \textrm{Tr}\big((\mathbf{U}^{t})^{T}\mathbf{R}_{k}\mathbf{U}^{t}\big)+\\
& \: \quad \quad  2 \textrm{Tr}\big((\mathbf{U}^{t})^{T}\mathbf{R}_{k}(\mathbf{U}-\mathbf{U}^{t})\big)\\
&=2\,\textrm{Tr}\big((\mathbf{U}^{t})^{T}\mathbf{R}_{k}\mathbf{U}\big)-\textrm{Tr}\big((\mathbf{U}^{t})^{T}\mathbf{R}_{k}\mathbf{U}^{t}\big) \triangleq g_{k}(\mathbf{U})
\end{array}
\end{equation*}
Then the surrogate problem is given by:
\begin{equation}
\begin{array}{ll}
{\max\limits_\mathbf{U}}   &   \min\limits_{k} \;\;{g_{k}(\mathbf{U})}\;\; \textrm{s.t.}\;\;  \begin{bmatrix}
         \mathbf{I}_{r} & \mathbf{U}^{T} \\ \mathbf{U} & \mathbf{I}_{n}  
    \end{bmatrix}\succcurlyeq 0\\
\end{array} \label{eq:func_fpca}
\end{equation}
Using the expression for $g_{k}(\mathbf{U})$ in (\ref{eq:func_fpca}) we get
\begin{equation}
\begin{array}{ll}
{\max\limits_\mathbf{U}}   &   \min\limits_{k} \;\;{2\,\textrm{Tr}(\mathbf{A}^{T}_{k}\mathbf{U})+c_{k}}\;\; \textrm{s.t.}\;\;  \begin{bmatrix}
         \mathbf{I}_{r} & \mathbf{U}^{T} \\ \mathbf{U} & \mathbf{I}_{n}  
    \end{bmatrix}\succcurlyeq 0\\
\end{array} \label{P3}
\end{equation}
where 
\begin{equation}
\mathbf{A}^{T}_{k}\triangleq(\mathbf{U}^{t})^{T}\mathbf{R}_{k}, \; c_{k}=-\textrm{Tr}((\mathbf{U}^{t})^{T}\mathbf{R}_{k}\mathbf{U}^{t}) \label{ak}
\end{equation}
Problem (\ref{P3}) is convex and can be reformulated as an SDP :
\begin{equation}
\begin{array}{ll}
{\max\limits_{\alpha,\mathbf{U}}}\;\;{\alpha}\;\; \\\textrm{s.t.}\;\; 2\,\textrm{Tr}(\mathbf{A}^{T}_{k}\mathbf{U})+c_{k} \geq \alpha\\\;\;\;\;\;\;\;\begin{bmatrix}
         \mathbf{I}_{r} & \mathbf{U}^{T} \\ \mathbf{U} & \mathbf{I}_{n}  
    \end{bmatrix}\succcurlyeq 0\\
\end{array} \label{P4}
\end{equation}
which can be solved using off-the-shelf solvers like CVX \cite{cvx}. The maximizer $\mathbf{U}$ of (\ref{P3}) (or equivalently (\ref{P4})) satisfies the constraint $\mathbf{U}^{T}\mathbf{U}=\mathbf{I}$ at each iteration of the algorithm (not only at convergence). This interesting property will be proved in what follows. In the process we will also derive a computationally simpler reformulation of (\ref{P4}). 

First, we note that the inner minimization problem in (\ref{P3}) can be reformulated using auxiliary variables $\boldsymbol{\mu} = \left[\mu_1, \cdots, \mu_K\right]^T$, as shown below:
\begin{equation}
\begin{array}{ll}
{\max\limits_\mathbf{U}}\,\,\min\limits_{\boldsymbol{\mu}} \;\;{\sum\limits_{k=1}^{K}\mu_{k}\textrm{g}_{k}(\mathbf{U})}\\ \textrm{s.t.}\;\; \mu_{k}\geq 0, \;\;\sum\limits_{k=1}^{K}\mu_{k}=1\\
\;\;\;\;\;\;\begin{bmatrix}
\mathbf{I}_{r} & \mathbf{U}^{T} \\ \mathbf{U} & \mathbf{I}_{n}  
\end{bmatrix}\succcurlyeq 0\\
\label{P5}
\end{array} 
\end{equation}
 The objective in (\ref{P5}) is a linear function of $\mathbf{U}$ for given $\boldsymbol{\mu}$, and it is linear in $\boldsymbol{\mu}$ for fixed $\mathbf{U}$. Futhermore the constraint sets for both $\mathbf{U}$ and $\boldsymbol{\mu}$ are compact and convex. Consequently, using the minimax theorem \cite{sion}, the max and min operators in (\ref{P5}) can be interchanged and we get the following equivalent problem:
 \begin{equation}
\begin{array}{ll}
{\min\limits_{\boldsymbol\mu}}\;\;\max\limits_{\mathbf{U}} \;\;{2\,\textrm{Tr}(\mathbf{A}^{T}\mathbf{U})+\sum\limits_{k=1}^{K}\mu_{k}c_{k}}\;\;\\
\textrm{s.t.}\;\;\mu_{k} \geq 0\;\;, \sum\limits_{k=1}^{K}\mu_{k}=1\;\;, \begin{bmatrix}
         \mathbf{I}_{r} & \mathbf{U}^{T} \\ \mathbf{U} & \mathbf{I}_{n}  
    \end{bmatrix}\succcurlyeq 0\\
\end{array} \label{P6}
\end{equation}
where
\begin{equation}
   \mathbf{A}\triangleq \sum\limits_{k=1}^{K}\mu_{k}\mathbf{A}^{T}_{k}
   \label{AD}
\end{equation}
The inner maximization problem in (\ref{P6}) can be solved in closed form. Consider the first term in the objective of (\ref{P6}). By Von-Neumann inequality \cite{von}, we have
\begin{equation*}
    \textrm{Tr}(\mathbf{A}^{T}\mathbf{U})\leq\sum\limits_{i=1}^{r}\sigma_{i}(\mathbf{A})\sigma_{i}(\mathbf{U})
\end{equation*}
where $\sigma_{i}(\mathbf{A})$ and $\sigma_{i}(\mathbf{U})$ denote the non-zero singular values of $\mathbf{A}$ and $\mathbf{U}$, respectively. Because $\sigma_{i}(\mathbf{U})\leq 1$, it follows that
\begin{equation*}
    \textrm{Tr}(\mathbf{A}^{T}\mathbf{U}) \leq \sum\limits_{i=1}^{r}\sigma_{i}(\mathbf{A}) 
\end{equation*}
with the equality attained for  
\begin{equation}
    \mathbf{U}^{*}=\mathbf{A}(\mathbf{A}^{T}\mathbf{A})^{-\frac{1}{2}}
    \label{OP}
\end{equation}
Indeed,
\begin{equation*}
\begin{array}{ll}
    \textrm{Tr}(\mathbf{A}^{T}\mathbf{U}^{*})=\textrm{Tr}((\mathbf{A}^{T}\mathbf{A})(\mathbf{A}^{T}\mathbf{A})^{-\frac{1}{2}})&=\textrm{Tr}((\mathbf{A}^{T}\mathbf{A})^{\frac{1}{2}})\\
    &=\sum\limits_{i=1}^{r}\sigma_{i}(\mathbf{A})
\end{array}
\end{equation*}
Note that $\mathbf{U}^{*}$ satisfies the constraint in (\ref{P1}) i.e. $(\mathbf{U}^{*})^{T}\mathbf{U}^{*}=\mathbf{I}$. Therefore, as claimed, the maximizer of (\ref{P4}) satisfies the constraint $\mathbf{U}^{T}\mathbf{U}=\mathbf{I}$ at each iteration. Now inserting (\ref{OP}) in (\ref{P6}) yields the following problem that remains to be solved:
\begin{equation}
\begin{array}{ll}
\min\limits_{\boldsymbol{\mu}} \;\;{2\,\sum\limits_{i=1}^{K}\sigma_{i}(\mathbf{A}(\boldsymbol{\mu}))+\sum\limits_{k=1}^{K}\mu_{k}c_{k}}\;\;\\
\textrm{s.t.}\;\;\mu_{k} \geq 0\;\;, \sum\limits_{k=1}^{K}\mu_{k}=1\;\;
\end{array} \label{P7}
\end{equation}
where we have stressed by notation that $\mathbf{A}$ is a function of $\boldsymbol{\mu}$. The first term in (\ref{P7}) is equal to 2 times the matrix nuclear norm of $\mathbf{A}(\boldsymbol{\mu})$, denoted $\|\mathbf{A}(\boldsymbol{\mu})\|_{*}$, and it is a convex function of $\mathbf{A}$ and hence $\left\{\mu_{k}\right\}$. Thus, (\ref{P7}) is a convex problem like (\ref{P4}) and can be reformulated as an SDP \cite{recht}. However compared to (\ref{P4}) the number of variables and constraints in (\ref{P7}) is smaller, which can be an advantage. In fact for $K=2$, (\ref{P7}) can be solved as a 1-dimensional problem via a bisection method. 


Once the minimizer $\boldsymbol{\mu}^{*}$ is obtained by solving (\ref{P7}), the corresponding $\mathbf{U}$ (which is also the maximizer of (\ref{P4})) can be obtained as:
\begin{equation}
    \mathbf{U}^{(t+1)}= \mathbf{A}(\boldsymbol{\mu}^{*})(\mathbf{A}^{T}(\boldsymbol{\mu}^{*})\mathbf{A}(\boldsymbol{\mu}^{*}))^{-\frac{1}{2}}
    \label{UD}
\end{equation}
 and it will serve as the next iterate. The MM-procedure will continue in the same manner till the iterates converge. The steps of the proposed algorithm are summarized in a pseudocode form in Algorithm 1.
\begin{algorithm}[ht!]
\caption{FPCA algorithm}
\fontsize{11}{13}\selectfont \textbf{Input}  Initial estimate $\mathbf{U}^{0}$, $\{\mathbf{R}_{k}\}_{k=1}^{K}$, and convergence threshold $\epsilon = 10^{-5}$.

\fontsize{11}{13}\selectfont Set $t=0$.

\fontsize{11}{13}\selectfont \hspace{0.1cm}\textbf{repeat}

\fontsize{11}{13}\selectfont \hspace{0.2cm}  Compute $\{\mathbf{A}_k,\;c_k\}$ in (\ref{ak}). 

\fontsize{11}{13}\selectfont \hspace{0.2cm}  Compute $\boldsymbol{\mu}^{*}$ by solving (\ref{P7}). 

\fontsize{11}{13}\selectfont \hspace{0.2cm} Obtain $\displaystyle \mathbf{U}^{t+1}$ from (\ref{UD}).

\fontsize{11}{13}\selectfont \hspace{0.2cm} Set $t=t+1$.

\fontsize{11}{13}\selectfont \hspace{0.1cm}\textbf{until}~$\displaystyle  \frac{\|\mathbf{U}^{t+1}-\mathbf{U}^{t}\|}{\|\mathbf{U}^t\|} \leq \epsilon $.

\fontsize{11}{13}\selectfont $\mathbf{U}_{\text{FPCA}}$ $=$ $\mathbf{U}^{t}$ at convergence.

\fontsize{11}{13}\selectfont \textbf{Output} ${\mathbf{U}_{\text{FPCA}}}$.
\label{alg:FPCA}
\end{algorithm}

The main burden of the proposed algorithm lies in the computation of $\{\mathbf{A}_{k}\}_{k=1}^{K}$, solving the SDP in (\ref{P7}) and computing $\mathbf{U}^{t+1}$ in (\ref{UD}). Computation of $\{\mathbf{A}_{k}\}_{k=1}^{K}$ involves evaluation of matrix-matrix products which can be done in $\mathcal{O}(Krn^{2})$ flops. Solving the SDP in (\ref{P7}) requires roughly $\mathcal{O}\big((n+r)^{4.5}\big)$ flops, and evaluating (\ref{UD}) requires $\mathcal{O}(nr^{2})$ + $\mathcal{O}(r^{3})$ flops, thus the total number of computations per iteration is on the order of $\mathcal{O}\big((n+r)^{4.5}\big)$ flops. The complexity of $\mathcal{O}\big((n+r)^{4.5}\big)$ seems to be on the higher side especially when compared to that of a recent state-of-the-art algorithm proposed in \cite{ami}, which has a complexity of $\mathcal{O}\big(n^{3}\big)$. Nonetheless, with modern day computers/workstations we could run Algorithm 1 for fairly larger dimensions such as $n=1000$, $r=100, K=100$. 

In the following, we propose an alternative approach  to solve (\ref{P7}) that helps reduce the computational burden. To do so, we start with the problem in (\ref{P7}):
\begin{equation}
\begin{array}{ll}
\min\limits_{\boldsymbol\mu} &{2\,\|\mathbf{A}(\boldsymbol{\mu})\|_{*}+\sum\limits_{k=1}^{K}\mu_{k}c_{k}}\\
\;\textrm{s.t.}&{\mu_{k} \geq 0\;\;, \sum\limits_{k=1}^{K}\mu_{k}=1}
\end{array} \label{P9}
\end{equation}
Let us introduce an auxillary variable $\boldsymbol{\Phi}$ ($\boldsymbol{\Phi} \succ \boldsymbol{0}$) and rewrite (\ref{P9}) in the form:
\begin{equation}
\begin{array}{ll}
\min\limits_{\boldsymbol{\mu,\Phi \succ 0}} &{\textrm{Tr}(\boldsymbol{\Phi}^{-1})+\textrm{Tr}(\mathbf{A}^{T}\mathbf{A}\boldsymbol{\Phi})+\sum\limits_{k=1}^{K}\mu_{k}c_{k}}\\
\;\;\;\textrm{s.t.}&{\mu_{k} \geq 0\;\;, \sum\limits_{k=1}^{K}\mu_{k}=1}
\end{array} \label{P10}
\end{equation}
The problems (\ref{P9}) and (\ref{P10}) are equivalent, which can be seen as follows. If we minimize (\ref{P10}) over $\boldsymbol{\Phi}$ for fixed $\boldsymbol{\mu}$, we get the minimizer $\boldsymbol{\Phi}^{*}=(\mathbf{A}^{T}\mathbf{A})^{-\frac{1}{2}}$ and substituting it back in (\ref{P10}) we get the problem in (\ref{P9}). Thus, instead of solving (\ref{P9}) one can solve (\ref{P10}) to obtain the minimizer of (\ref{P9}). The problem in (\ref{P10}) can be solved using an alternating minimization approach : for given $\boldsymbol{\mu}$, say $\boldsymbol{\mu}^{t}$, we get the minimizer $\boldsymbol{\Phi}^{t}$ as explained above. For fixed $\boldsymbol{\Phi}$ $=$ $\boldsymbol{\Phi}^{t},$ the minimizer $\boldsymbol{\mu}$ can be obtained by solving the following convex problem:
\begin{equation}
\begin{array}{ll}
\min\limits_{\boldsymbol\mu} &{\textrm{Tr}(\mathbf{A}^{T}\mathbf{A}(\boldsymbol{\Phi}^{t}))+\sum\limits_{k=1}^{K}\mu_{k}c_{k}}\\
\textrm{s.t.} &\mu_{k} \geq 0\;\;, \sum\limits_{k=1}^{K}\mu_{k}=1
\end{array} \label{P11}
\end{equation}
which can be reformulated as a quadratic program (QP):
\begin{equation}
\begin{array}{ll}
\min\limits_{\boldsymbol\mu} & \boldsymbol{\mu}^T \mathbf{Q} \boldsymbol{\mu} +\sum\limits_{k=1}^{K}\mu_{k}c_{k}\\
\textrm{s.t.} &\mu_{k} \geq 0\;\;, \sum\limits_{k=1}^{K}\mu_{k}=1
\end{array} \label{QP}
\end{equation}
where $[\mathbf{Q}]_{i,j} \triangleq \text{Tr}(\mathbf{A}_i^T \boldsymbol{\Phi}^t \mathbf{A}_j)$. The QP in (\ref{QP}) can be solved using standard solvers whose computational complexity including the update of 
$\boldsymbol{\Phi}^t$ is on the order of $\mathcal{O}(K^{3}) + \mathcal{O}(n^3)$ flops. This is much less than $\mathcal{O}((n+r)^{4.5})$ but note that, unlike the problem in (\ref{P7}), the problem in (\ref{QP}) has to be solved multiple times (for different $\boldsymbol{\Phi}^{t}$ matrices); fortunately usually only a few iterations are needed (typically $<10$). The iterative steps of the above alternating minimization algorithm for computing the solution $\boldsymbol{\mu}^*$ to (\ref{P7}) are summarized in Algorithm 2. 
\begin{algorithm}[ht!]
\caption{Alternating minimization approach for solving (\ref{P7})}
\fontsize{11}{13}\selectfont \textbf{Input}  Initial estimate $\boldsymbol{\mu}^{0}$, \{$c_{k}$, $\mathbf{A}^{T}_{k}$\} and convergence threshold $\epsilon = 10^{-5}$.

\fontsize{11}{13} \selectfont Set $t=0$.

\fontsize{11}{13}\selectfont \hspace{0.1cm}\textbf{repeat}

\fontsize{11}{13}\selectfont \hspace{0.2cm}  Compute $\displaystyle \boldsymbol{\Phi}^{t}=\big(\mathbf{A}^{T}(\boldsymbol{\mu}^{t})\mathbf{A}(\boldsymbol{\mu}^{t})\big)^{-\frac{1}{2}}$. 

\fontsize{11}{13}\selectfont \hspace{0.2cm} Obtain $\boldsymbol{\mu}^{t+1}$ by solving (\ref{QP}).

\fontsize{11}{13}\selectfont \hspace{0.2cm} Set $t = t+1$.

\fontsize{11}{13}\selectfont \hspace{0.1cm}\textbf{until}~$ \displaystyle \frac{\|\boldsymbol{\mu}^{t+1}-\boldsymbol{\mu}^{t}\|}{\|\boldsymbol{\mu}^t\|} \leq \epsilon $.

\fontsize{11}{13}\selectfont $\boldsymbol{\mu}^{*}=\boldsymbol{\mu}^{t}$ at convergence.

\fontsize{11}{13}\selectfont \textbf{Output} $\boldsymbol{\mu}^{*}$.
\end{algorithm}
\section{FSPCA and FRPCA}
\subsection{Fair Sparse PCA}
In this section we present an extension of the proposed method which can handle a penalty that promotes the sparsity of the fair principal components. For convenience we repeat here the statement of the sparse FPCA problem (see (\ref{eq:fpca_2})):
\begin{equation}
\begin{array}{ll}
{\max\limits_\mathbf{U}}\;\;\min\limits_{k\in [1,K]} \;\;\{f_{k}(\mathbf{U})-\lambda \|\mathbf{U}\|_{1}\}\\\;\;\textrm{s.t.}\;\;\;\;\;\mathbf{U}^{T}\mathbf{U}=\mathbf{I}\\
\end{array} \label{eq:fpca_sparse}
\end{equation}
where $\lambda$ is a pre-specified factor that controls the sparsity of $\mathbf{U}$. Similarly to the FPCA case, we relax the semi-orthogonality constraint:
\begin{equation}
\begin{array}{ll}
{\max\limits_\mathbf{U}}&\min\limits_{k\in [1,K]} \;\;f_{k}(\mathbf{U})-\lambda \|\mathbf{U}\|_{1}\\\textrm{s.t.}&\mathbf{U}^{T}\mathbf{U} \preccurlyeq\mathbf{I} \implies\begin{bmatrix}
         \mathbf{I}_{r} & \mathbf{U}^{T} \\ \mathbf{U} & \mathbf{I}_{n}  
    \end{bmatrix}\succcurlyeq 0 \\
\end{array} \label{eq:fpca_sparse2}
\end{equation}
For fixed $\mathbf{U}=\mathbf{U}^{t}$, we can minorize the quadratic functions $\{f_{k}(\mathbf{U})\}$ using their tangent hyperplanes, and obtain the following surrogate problem
\begin{equation}
\begin{array}{ll}
{\max\limits_\mathbf{U}}\min\limits_{k} &\{g_{k}(\mathbf{U})-\lambda \|\mathbf{U}\|_{1}\}\\\textrm{s.t.}&\begin{bmatrix}
         \mathbf{I}_{r} & \mathbf{U}^{T} \\ \mathbf{U} & \mathbf{I}_{n}  
    \end{bmatrix}\succcurlyeq 0 \\
\end{array} \label{S1}
\end{equation}
where as before $g_{k}(\mathbf{U})=2\,\textrm{Tr}(\mathbf{A}_{k}^{T}\mathbf{U})+c_{k}$. The above problem is convex (more exactly an SDP) and can be solved via CVX. Similar to the FPCA case, the maximizer of (\ref{S1}) can be shown to satisfy the constraint $\mathbf{U}^{T}\mathbf{U}=\mathbf{I}$. To do so, we reformulate the problem in  (\ref{S1}) using auxiliary variables $\mathbf{B}$ and $\boldsymbol{\mu}$ as follows:
\begin{equation}
\begin{array}{ll}
{\max\limits_{\mathbf{U}}}\;\;\min\limits_{\boldsymbol{\mu},\mathbf{B}} \;\;{\sum\limits_{k=1}^{K}\mu_{k}g_{k}(\mathbf{U})+\lambda\,\textrm{Tr}(\mathbf{B}^{T}\mathbf{U})}\;\;\\
\textrm{s.t.}\;\;\mu_{k} \geq 0\;\;, \sum\limits_{k=1}^{K}\mu_{k}=1\;\;, \begin{bmatrix}
         \mathbf{I}_{r} & \mathbf{U}^{T} \\ \mathbf{U} & \mathbf{I}_{n}  
    \end{bmatrix}\succcurlyeq 0,|[\mathbf{B}]_{i,j}| \leq 1 \; \forall i,j
\end{array} \label{S2}
\end{equation}
Minimizing (\ref{S2}) over $\mathbf{B}$ and $\boldsymbol{\mu}$ we get the objective in the problem (\ref{S1}). Thus (\ref{S1}) and (\ref{S2}) are equivalent problems. Using the minimax theorem (the objective and the constraints of (\ref{S2}) satisfy the required conditions), the max and min operators can be swapped:
\begin{equation}
\begin{array}{ll}
\min\limits_{\boldsymbol{\mu},\mathbf{B}}\;\; {\max\limits_{\mathbf{U}}} \;\;{\sum\limits_{k=1}^{K}\mu_{k}g_{k}(\mathbf{U})+\lambda\,\textrm{Tr}(\mathbf{B}^{T}\mathbf{U})}\;\;\\
\textrm{s.t.}\;\;\mu_{k} \geq 0\;\;, \sum\limits_{k=1}^{K}\mu_{k}=1\;\;, \begin{bmatrix}
         \mathbf{I}_{r} & \mathbf{U}^{T} \\ \mathbf{U} & \mathbf{I}_{n}  
    \end{bmatrix}\succcurlyeq 0,|[\mathbf{B}]_{i,j}| \leq 1 \; \forall i,j
\end{array} \label{S3}
\end{equation}
Using the expression for $g_{k}(\mathbf{U})=2\,\textrm{Tr}(\mathbf{A}_{k}^{T}\mathbf{U})+c_{k}$, (\ref{S3}) can be rewritten as:
\begin{equation}
\begin{array}{ll}
\min\limits_{\boldsymbol{\mu},\mathbf{B}}\;\; {\max\limits_{\mathbf{U}}} \;\;{2\,\textrm{Tr}((\mathbf{A}+\displaystyle \frac{\lambda}{2}\mathbf{B})^{T}\mathbf{U})+\sum\limits_{k=1}^{K}\mu_{k}c_{k}}\;\;\\
\textrm{s.t.}\;\;\mu_{k} \geq 0\;\;, \sum\limits_{k=1}^{K}\mu_{k}=1\;\;, \begin{bmatrix}
         \mathbf{I}_{r} & \mathbf{U}^{T} \\ \mathbf{U} & \mathbf{I}_{n}  
    \end{bmatrix}\succcurlyeq 0,\;|[\mathbf{B}]_{i,j}| \leq 1 \; \forall i,j
\end{array} \label{eqB}
\end{equation}
Similar to (\ref{OP}), the maximizer $\mathbf{U}$ of (\ref{eqB}) can be obtained in closed form:
\begin{equation*}
    \mathbf{U}^{*}=(\mathbf{A}+\frac{\lambda}{2}\mathbf{B})\big((\mathbf{A}+\frac{\lambda}{2}\mathbf{B})^{T}(\mathbf{A}+\frac{\lambda}{2}\mathbf{B})\big)^{-\frac{1}{2}}
\end{equation*}
which satisfies the constraint $\mathbf{U}^{T}\mathbf{U}=\mathbf{I}$. Thus, also in the FSPCA case, the MM iterates satisfy the semi-orthogonality constraint. The pseudocode of FSPCA is summarized in Algorithm 3.
\begin{algorithm}[h!]
\caption{FSPCA algorithm}
\fontsize{11}{13}\selectfont \textbf{Input}  Initial estimate $\mathbf{U}^{0}$, $\{\mathbf{R}_{k}\}$, $\lambda$, and convergence threshold $\epsilon = 10^{-5}$.

\fontsize{11}{13} \selectfont Set $t=0$.

\fontsize{11}{13}\selectfont \hspace{0.1cm}\textbf{repeat}

\fontsize{11}{13}\selectfont \hspace{0.2cm}  Compute \{$\mathbf{A}_{k}$,\; $c_{k}$\} in (\ref{ak}).

\fontsize{11}{13}\selectfont \hspace{0.2cm} Obtain $\mathbf{U}^{t+1}$ by solving (\ref{S1}).

\fontsize{11}{13}\selectfont \hspace{0.2cm} Set $t = t+1$. 

\fontsize{11}{13}\selectfont \hspace{0.1cm}\textbf{until}~$ \displaystyle \frac{\|\mathbf{U}^{t+1}-\mathbf{U}^{t}\|}{\|\mathbf{U}^t\|} \leq \epsilon $.

\fontsize{11}{13}\selectfont $\mathbf{U}_{\text{SFPCA}}=\mathbf{U}^{t}$ at convergence.

\fontsize{11}{13}\selectfont \textbf{Output} $\mathbf{U}_{\text{SFPCA}}$.
\end{algorithm}

\subsection{Fair Robust PCA}
We next discuss how the proposed approach can be modified to solve the fair robust PCA formulation stated in (\ref{eq:fpca_robust}):
\begin{equation*}
\begin{array}{ll}
{\max\limits_\mathbf{U}} & \min\limits_{k\in [1,K]} \;\;\ \|\mathbf{U}^{T}\mathbf{Y}_{k}\|_{1}\;\; \textrm{s.t.}\;\; \mathbf{U}^{T}\mathbf{U}=\mathbf{I}\\
\end{array}
\end{equation*}
Similar to what we have done before, we relax the semi-orthogonality constraint:
\begin{equation}
\begin{array}{ll}
{\max\limits_\mathbf{U}} \;\; \min\limits_{k\in [1,K]} \;\;\ \|\mathbf{U}^{T}\mathbf{Y}_{k}\|_{1}\\\;\;\textrm{s.t.}\;\;\;\begin{bmatrix}
         \mathbf{I}_{r} & \mathbf{U}^{T} \\ \mathbf{U} & \mathbf{I}_{n}  
    \end{bmatrix}\succcurlyeq 0\\
\end{array} 
\label{L1}
\end{equation}
Compared to the previous cases, we have a different data fitting function here. Nonetheless, $\|\mathbf{U}^{T}\mathbf{Y}_{k}\|_{1}$ can be minorized  as follows. For a given $\mathbf{U}^{t}$,
\begin{equation}
\begin{array}{ll}
    \|\mathbf{U}^{T}\mathbf{Y}_{k}\|_{1}&=\sum\limits_{i,j}|(\mathbf{U}^{T}\mathbf{Y}_{k})_{ij}|\geq \sum\limits_{i,j}(\mathbf{U}^{T}\mathbf{Y}_{k})_{ij}\frac{((\mathbf{U}^{t})^{T}\mathbf{Y}_{k})_{ij}}{|((\mathbf{U}^{t})^{T}\mathbf{Y}_{k})_{ij}|}\\
&\quad\quad \quad \quad \quad =\sum\limits_{i,j}(\mathbf{U}^{T}\mathbf{Y}_{k})_{ij}\text{sgn}(((\mathbf{U}^{t})^{T}\mathbf{Y}_{k})_{ij})\\
&\quad \quad \quad \quad \quad =\textrm{Tr}(\mathbf{U}^{T}\mathbf{Y}_{k}\mathbf{W}_k)
\end{array}
\end{equation}
where
\begin{equation}
  \mathbf{W}_k \triangleq \text{sgn}((\mathbf{U}^{t})^{T}\mathbf{Y}_{k})
  \label{W}
\end{equation}

Thus the surrogate problem corresponding to (\ref{L1}) is given by:
\begin{equation}
\begin{array}{ll}
{\max\limits_\mathbf{U}} \;\; \min\limits_{k} \;\;\ \textrm{Tr}(\mathbf{U}^{T}\mathbf{Y}_{k}\mathbf{W}_k)\\\textrm{s.t.}\;\;\;\begin{bmatrix}
         \mathbf{I}_{r} & \mathbf{U}^{T} \\ \mathbf{U} & \mathbf{I}_{n}  
    \end{bmatrix}\succcurlyeq 0\\
\end{array} \label{RPCA}
\end{equation}
which is once again an SDP (after rewriting in epigraph form) that can be solved via CVX. Using a similar argument to that employed in the cases of FPCA and FSPCA, here too one can show that the solution of (\ref{RPCA}) satisfies the semi-orthogonality constraint. The pseudocode of the FRPCA is summarized in Algorithm 4.
\begin{algorithm}[h!]
\caption{FRPCA algorithm}
\fontsize{11}{13}\selectfont \textbf{Input}  Initial estimate $\mathbf{U}^{0}$, $\{\mathbf{R}_{k}\}$, $\lambda$, and convergence threshold $\epsilon = 10^{-5}$.

\fontsize{11}{13} \selectfont Set $t=0$.

\fontsize{11}{13}\selectfont \hspace{0.1cm}\textbf{repeat}

\fontsize{11}{13}\selectfont \hspace{0.2cm}  Compute $\{\mathbf{W}_k\}$ in (\ref{W}).

\fontsize{11}{13}\selectfont \hspace{0.2cm} Compute $\mathbf{U}^{t+1}$ by solving (\ref{RPCA}). 

\fontsize{11}{13}\selectfont \hspace{0.2cm} Set $t = t+1$. 

\fontsize{11}{13}\selectfont \hspace{0.2cm}\textbf{until}~$\displaystyle \frac{\|\mathbf{U}^{t+1}-\mathbf{U}^{t}\|}{\|\mathbf{U}^t\|} \leq \epsilon $.

\fontsize{11}{13}\selectfont $\mathbf{U}_{\text{FRPCA}}$ = $\mathbf{U}^{t}$ at convergence.

\fontsize{11}{13}\selectfont \textbf{Output} $\mathbf{U}_{\text{FRPCA}}$.
\end{algorithm}

\text{We conclude this section with a number of remarks on the}\\
\text{proposed algorithms.}
\begin{itemize}
    \item The value of $\lambda$ in the case of FSPCA and the rank $r$ for all the algorithms are assumed to be prespecified. When $\lambda$ and $r$ are not specified they can be chosen using model selection approaches or cross validation.  
    \item In the case of FSPCA and FRPCA, although we did not prove that we can relax the semi-orthogonality constraint in the original problem formulations (like we did for FPCA), we showed that the corresponding surrogate problems tightly satisfy the semi-orthogonality constraint, which means that their solutions are feasible for the original problems. 
    \item If desired a sparsity inducing penalty can also be added to FRPCA problem and an MM algorithm can be developed.
    \item In the case of FSPCA and FRPCA, QPs similar to (\ref{QP}) can be derived and solved instead of solving the SDPs in (\ref{S1}) and (\ref{RPCA}).
    \item Regarding the convergence of the proposed methods, by the properties of MM the objective function is monotonically increasing at each iteration. Then, because the cost functions of FPCA, FSPCA and FRPCA are bounded above, the convergence of the proposed algorithms follows from the general results proved in \cite{MM}. 
    \item Regarding the initialization ($\mathbf{U}^0$) for the proposed methods, in the case of FPCA and FSPCA we observed in the numerical experiments that the objective functions were often unimodal. So, we have used random semi-orthogonal matrices to initialize the algorithms. On the other hand, the objective function of FRPCA appeared to be multi-modal (see the discussion in Section V.C), thus we have initialized the FRPCA algorithm with the FPCA solution. 
    \item Finally we note that RPCA and SPCA are important problems in their own right and a multitude of algorithms have been suggested for solving them (see, e.g. \cite{RPCA, sparsePCA1, sparsePCA2, sparsePCA3}). Our algorithms FRPCA and FPSCA with $K=1$ can also be used to solve the RPCA and SPCA problems.
\end{itemize}
\begin{figure}[ht]
\begin{centering}
\begin{tabular}{c}
\includegraphics[width=8cm,height=6cm]{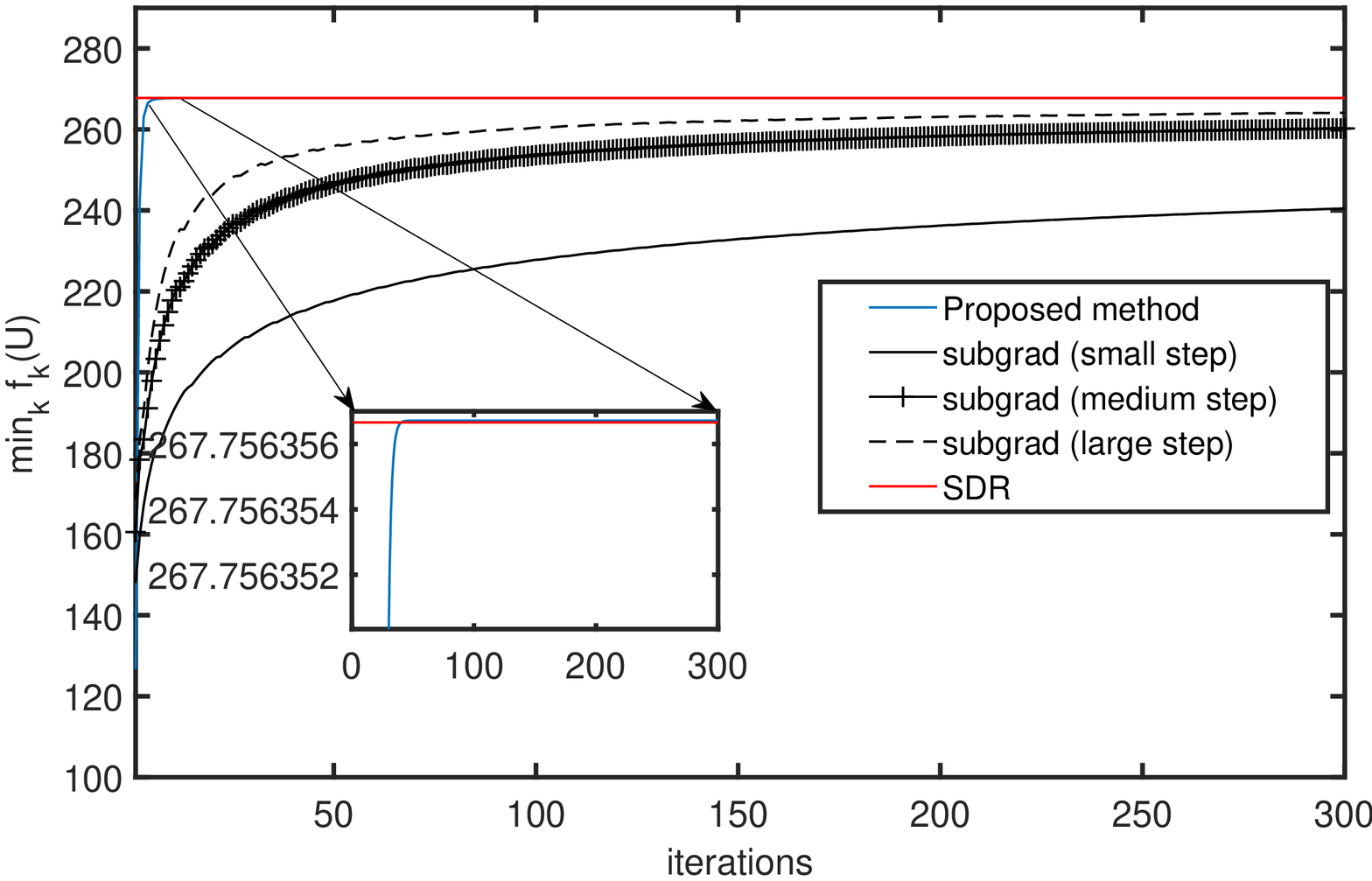} \tabularnewline
a) $K=2$\tabularnewline
\includegraphics[width=8cm,height=6cm]{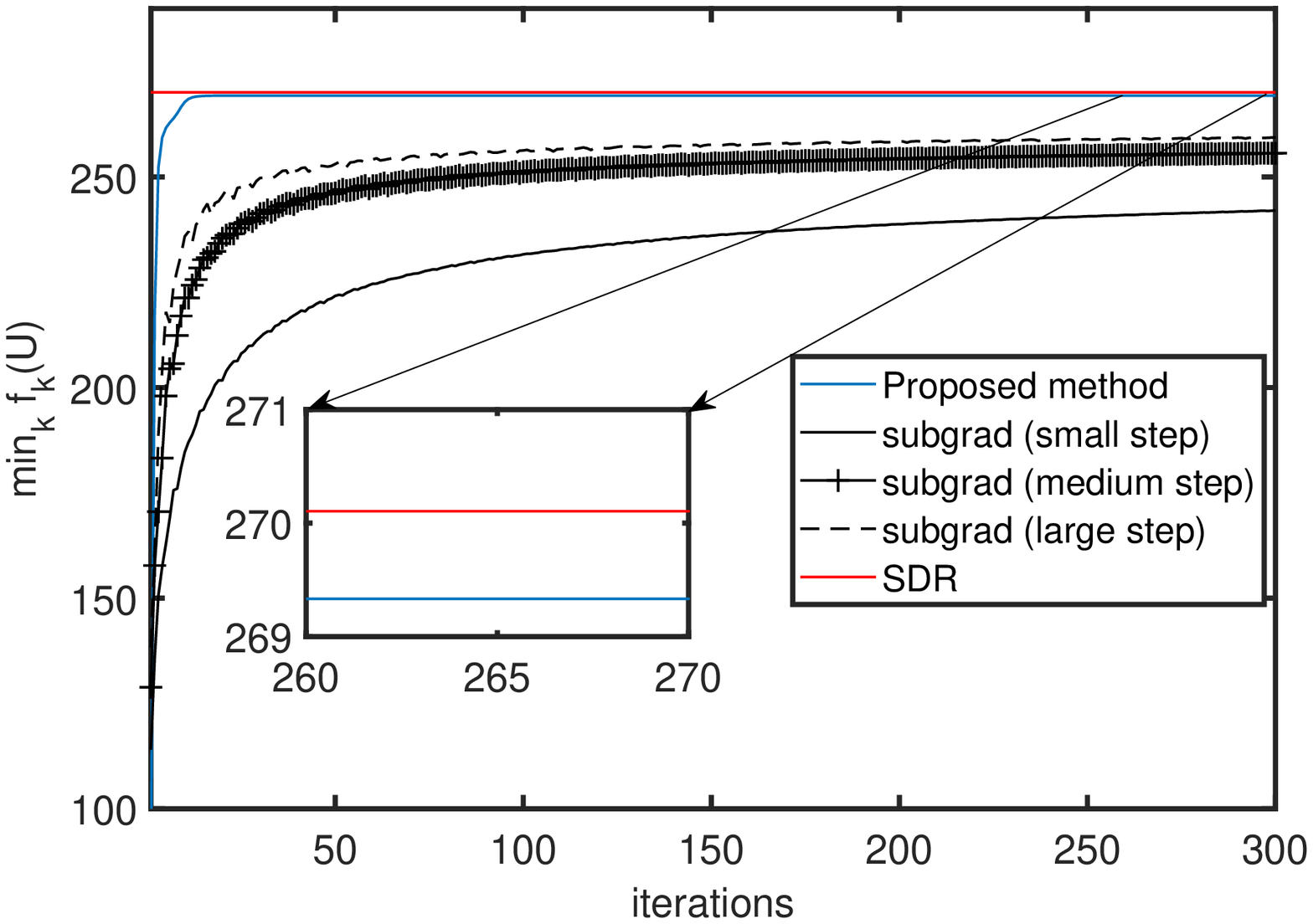}\tabularnewline
b)  $K=5$\tabularnewline
\tabularnewline
\end{tabular}
\par\end{centering}
\caption{Synthetic data set: Objective vs iteration plots for $n=10$, $N=100$, and $r=4$. In the case of SDR, for the simulation settings in figure 1b, the rank of optimal $\mathbf{P}$ was found to be $6$.}
\label{fig:1}
\end{figure}
\begin{figure}[ht]
\begin{centering}
\begin{tabular}{c}
\includegraphics[width=8cm,height=6cm]{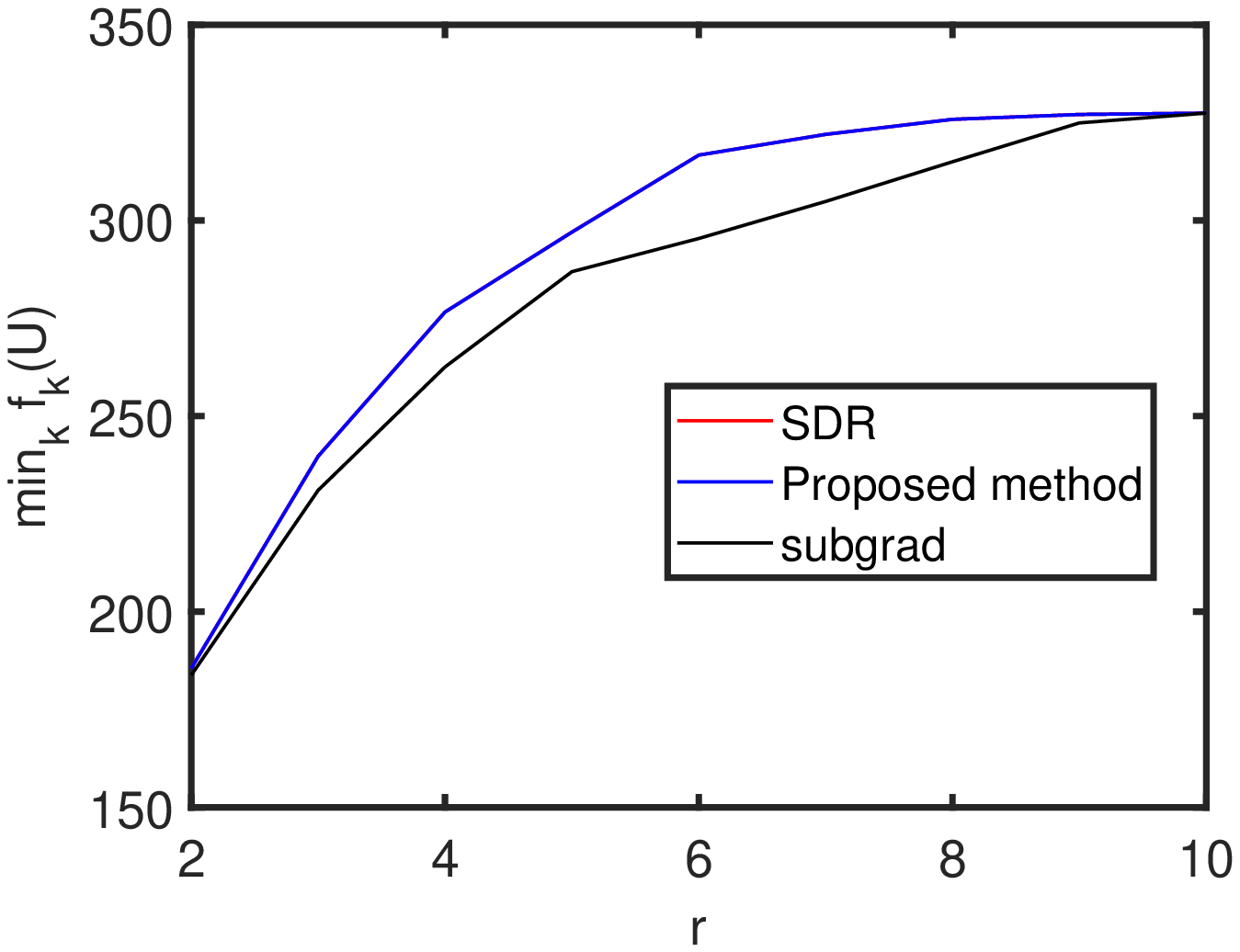} \tabularnewline
a) $K=2$ \tabularnewline
\includegraphics[width=8cm,height=6cm]{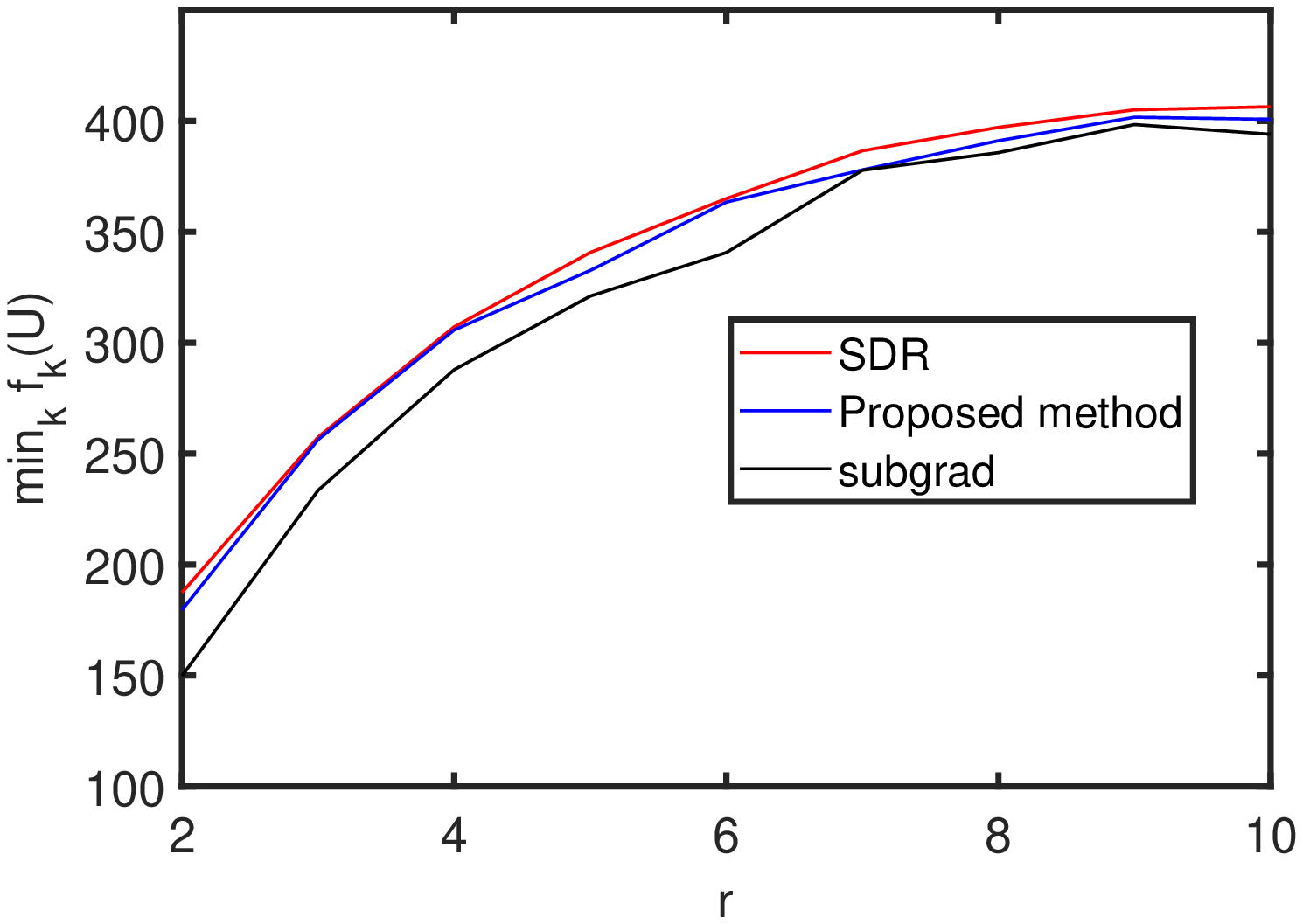}\tabularnewline
 b) $K=5$ 
\tabularnewline
\end{tabular}
\par\end{centering}
\caption{Synthetic data : Objective vs $r$ plots for $n=10$, $N=100$.}
\label{fig:2}
\end{figure}

\section{Numerical simulation results}
In this section, we will compare the performance of the proposed FPCA method with the SDR method of \cite{samadi1} and the sub-gradient method from \cite{ami}. The SDR approach uses the parametrization $\mathbf{P} = \mathbf{U} \mathbf{U}^T$ and relaxes the rank$(\mathbf{P})=r$ constraint to $\text{Tr}(\mathbf{P})=r$ and the semi-orthogonality constraint to $\boldsymbol{0} \preccurlyeq$ $\mathbf{P} \preccurlyeq \mathbf{I}$. For $K=2$, the aforementioned relaxation is tight but for $K\geq3$ this is not necessarily true and the method often yields solutions with rank$(\mathbf{P})>r$. Being a relaxation the objective value attained by SDR is an upper bound which is not always achievable. We first perform simulations based on synthetic data sets and later include simulation results for a real-life data set. We will also present numerical simulation results showing the performance of the proposed FSPCA and FRPCA methods.

\subsection{Synthetic data}
The data samples for any class $k$ is generated as $\{\mathbf{y}_i = \mathbf{R}_k^{\frac{1}{2}}\mathbf{e}_i \}$, where $\mathbf{R}_k^{\frac{1}{2}}$ was randomly generated, and the elements of $\mathbf{e}_i$ are i.i.d. random variables sampled from a Gaussian distribution with zero mean and variance one. Let $n=10, N=100,\,\text{and}\;\,r = 4$, and each class has the same number of data samples equal to $\frac{N}{K}$. We first show the objective vs. iteration plots for the three methods in Figure 1 for two different values of $K$. It can be seen from this figure that the proposed FPCA method converges quickly to the SDR upper bound (see the zoom insert in Figure 1a), while the method of \cite{ami} converges slowly to a suboptimal value. For $K>2$, the SDR upper bound may not be achievable, see for example Figure 1b (zoom insert). For fairness, we ran the method of \cite{ami} for different choices of the step length (small, medium and large ) and from Figure 1 one can see that even for different step-lengths the method still converges to suboptimal values. In Figures 2a and 2b, we show the objective values obtained by the three methods for different values of $r$ and $K$. From these figures it can be seen that the proposed FPCA approach always finds a better objective value than the method from \cite{ami} and reaches the SDR bound whenever the latter is achievable.

\subsection{Credit data}
In this simulation we evaluate the performance of the methods on a credit data set from \cite{samadi1}. First the data is mean centered. The parameters of the credit data are $n=21$, $K=2$, $N=30000$ (with class $1$ having $5385$ data samples and class $2$ having $24615$ data samples). In Figure 3a, we plot the objective vs iteration for the proposed FPCA method and the sub-gradient method of \cite{ami}. The proposed method quickly converges to the SDR bound. On the other hand, even for a careful choice of the step size, the method of \cite{ami} takes nearly a million iterations to converge to a suboptimal value. In Figure 3b we show the optimal objective values obtained by the three methods for $r$ in the interval $[1,20]$. It can be seen from this figure that the FPCA method reaches the same objective as SDR.
\begin{figure}[ht]
\begin{centering}
\begin{tabular}{cc}
\includegraphics[width=8cm,height=6cm]{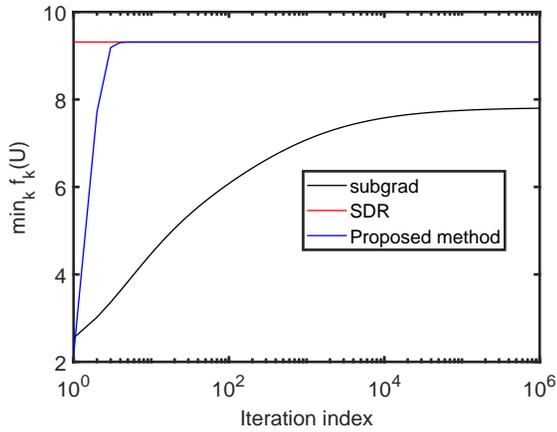}\tabularnewline
a) Objective vs iteration ($r=2$) \tabularnewline 
\includegraphics[width=8cm,height=6cm]{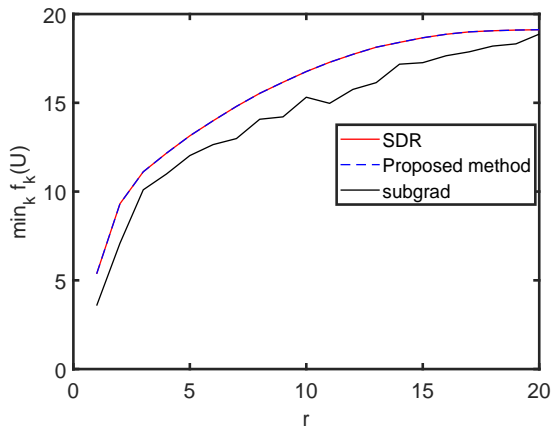}\tabularnewline
 b) Objective vs $r$ 
\tabularnewline
\end{tabular}
\par\end{centering}
\caption{Credit data set with $n=21$, $N=30000$, $K=2$.}
\label{fig:3}
\end{figure}

\subsection{Fair robust PCA}
In this simulation, we generate the data as in Section V.A but now some data samples for randomly chosen classes are corrupted by outliers. The outliers, which are roughly $10\%$ of the total number of samples, are generated from a Gaussian distribution with mean $\alpha$ and variance one. In Figure 4a we plot the objective vs iteration for the proposed FPCA algorithm for ten different initializations. Unlike the FPCA and the FSPCA case, the objective in (\ref{eq:fpca_robust}) has several local maxima, which is evident from the plot as the method converges to different local maxima depending on the intialization. In Figure 4b, we vary $\alpha$ in the interval $[0,20]$ and compute the average normalized error between the estimated subspaces (calculated using 100 Monte-Carlo runs):
\begin{equation}
\begin{array}{ll}
\text{Normalized subspace error} = \frac{\|\hat{\mathbf{U}}\hat{\mathbf{U}}^T-\tilde{\mathbf{U}}_{\text{FPCA}}\tilde{\mathbf{U}}_{\text{FPCA}}^T\|}{\|\tilde{\mathbf{U}}_{\text{FPCA}}\tilde{\mathbf{U}}_{\text{FPCA}}^T\|}
\end{array} \label{error}
\end{equation}
where $\tilde{\mathbf{U}}_{\text{FPCA}}$ denotes the FPCA estimate of the principal components obtained from outlier-free data, and $\hat{\mathbf{U}}$ denotes either $\mathbf{U}_{\text{FPCA}}$ or $\mathbf{U}_{\text{FRPCA}}$ obtained in the presence of outliers. It can be seen from the figure that the FRPCA performs well in the presence of the outliers and has comparatively low errors even in the presence of large outliers. 

\begin{figure}[ht]
\begin{centering}
\begin{tabular}{c}
\includegraphics[width=8cm,height=6cm]{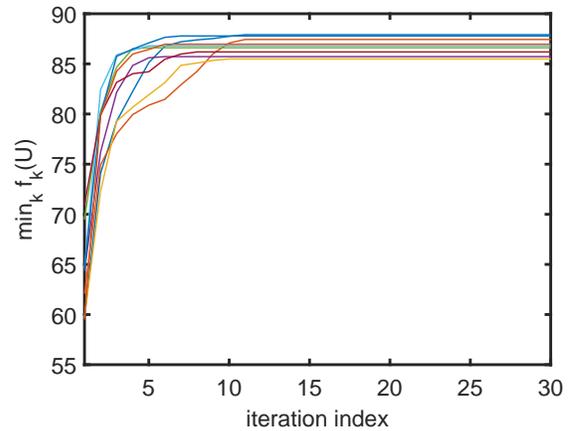} \tabularnewline 
a) Objective vs iteration (for 10 random initializations) \tabularnewline
\includegraphics[width=8cm,height=6cm]{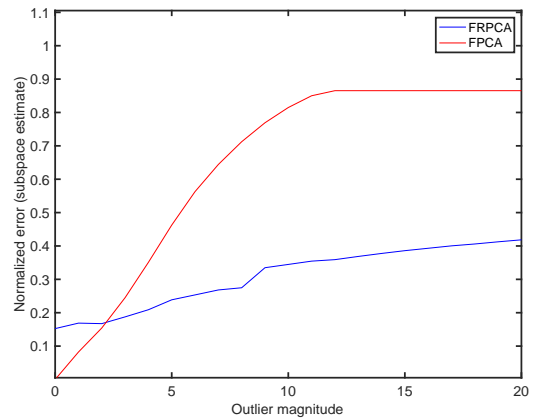}\tabularnewline
b) Normalized subspace error vs Outlier magnitude ($\alpha$).  
\tabularnewline
\end{tabular}
\par\end{centering}
\caption{Performance of FRPCA algorithm for $n=10$, $N=100$, $K=2$, $r=4$ and $10 \%$ outliers.}
\label{fig:4}
\end{figure}

\begin{figure}[ht]
\begin{centering}
\begin{tabular}{c}
\includegraphics[width=8cm,height=6cm]{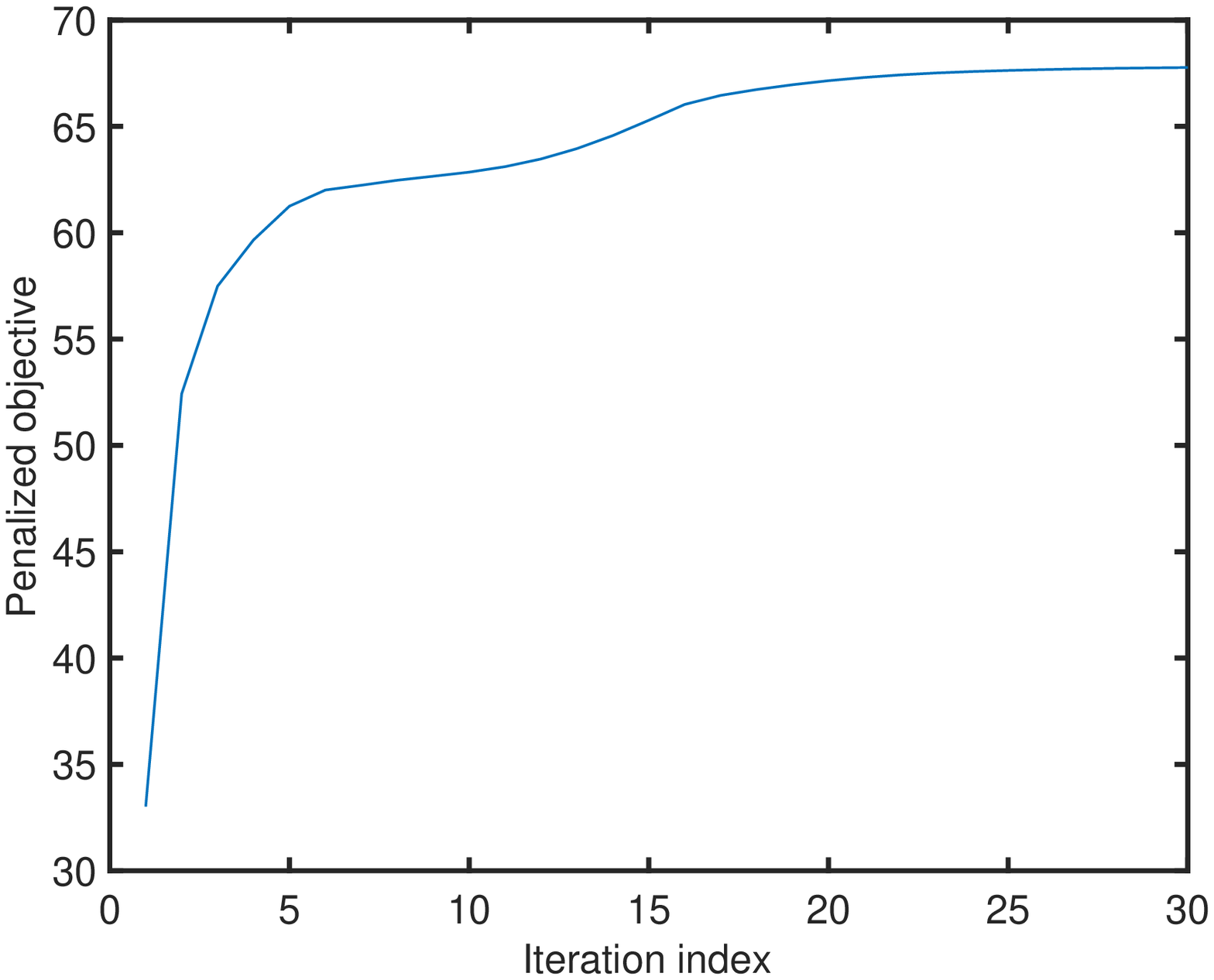} \\
a) The Objective in \eqref{eq:fpca_sparse} vs iteration \\
\includegraphics[width=8cm,height=6cm]{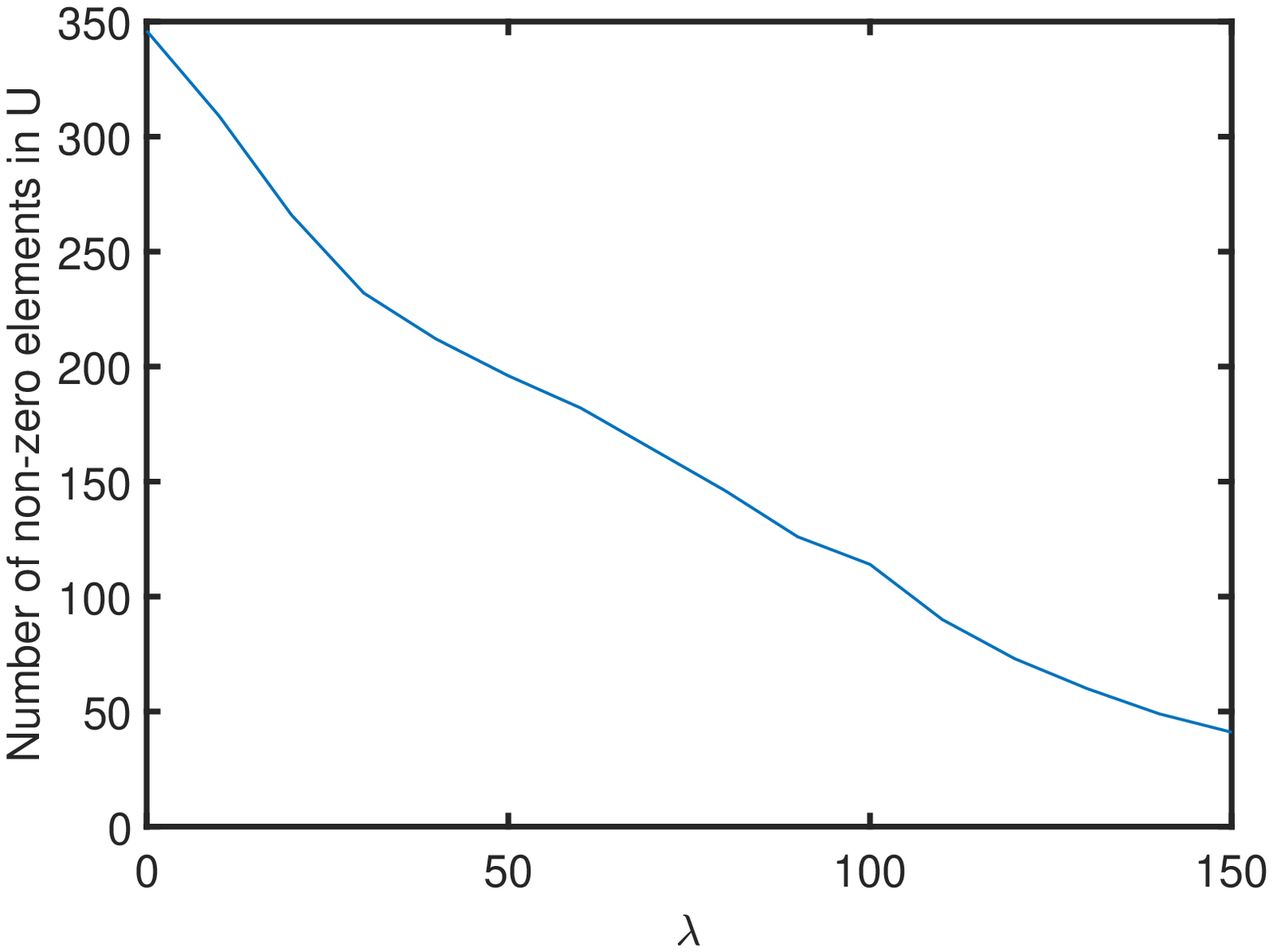} \\ 
b) Number of non-zero elements vs $\lambda$
\tabularnewline
\end{tabular}
\par\end{centering}
\caption{FSPCA: $n=40, \;r=10,\; N=200, \; K= 2$.}

\end{figure}

\subsection{Fair sparse PCA}
 In this simulation we generate the data as in Section V.A with the choice of parameters: $n=40, r=10, N=200, K=2$. In Figure 5a, we show the evolution of the objective in (\ref{eq:fpca_sparse}) vs iterations for $\lambda=5$. The proposed algorithm monotonically increases the penalized objective as expected. The choice of $\lambda$ determines the sparsness of $\mathbf{U}$. In Figure 5b we show the number of non-zero elements in $\mathbf{U}$ vs $\lambda$ (as expected) when $\lambda$ increases the sparsity of the estimated $\mathbf{U}$ increases.

\section{Conclusion}
In this paper we have proposed a new MM algorithm for solving the fair PCA (FPCA) problem. The proposed algorithm is computationally efficient and monotonically increases the FPCA objective at each iteration. Furthermore, the proposed FPCA algorithm does not require the selection of any hyperparameter. We have also proposed two MM algorithms to solve the fair robust PCA and fair sparse PCA problems. Finally, we performed numerical simulations on both synthetic data sets and a real-life data set and compared the performance of the proposed approach with that of two state-of-the-art approaches.

\bibliographystyle{ieeetr}
\bibliography{ref}

\begin{thebibliography}{10}

\bibitem{samadi1}
S.~Samadi, U.~Tantipongpipat, J.~H. Morgenstern, M.~Singh, and S.~Vempala,
  ``{The price of fair PCA: One extra dimension},'' {\em Advances in neural
  information processing systems}, vol.~31, 2018.

\bibitem{PCA1}
I.~T. Jolliffe, {\em Principal component analysis for special types of data}.
\newblock Springer, 2002.

\bibitem{PCAimage1}
C.~Clausen and H.~Wechsler, ``{Color image compression using PCA and
  backpropagation learning},'' {\em Pattern Recognition}, vol.~33, no.~9,
  pp.~1555--1560, 2000.

\bibitem{PCAimage2}
Q.~Du and J.~E. Fowler, ``Low-complexity principal component analysis for
  hyperspectral image compression,'' {\em The International Journal of High
  Performance Computing Applications}, vol.~22, no.~4, pp.~438--448, 2008.

\bibitem{PCASP1}
F.~Castells, P.~Laguna, L.~S{\"o}rnmo, A.~Bollmann, and J.~M. Roig,
  ``{Principal component analysis in ECG signal processing},'' {\em EURASIP
  Journal on Advances in Signal Processing}, vol.~2007, pp.~1--21, 2007.

\bibitem{PCASP2}
A.~Subasi and M.~I. Gursoy, ``{EEG signal classification using PCA, ICA, LDA
  and support vector machines},'' {\em Expert systems with applications},
  vol.~37, no.~12, pp.~8659--8666, 2010.

\bibitem{PCAfinance}
R.~S. Tsay, {\em Analysis of financial time series}.
\newblock John wiley \& sons, 2005.

\bibitem{PCAbio1}
A.~Giuliani, ``The application of principal component analysis to drug
  discovery and biomedical data,'' {\em Drug discovery today}, vol.~22, no.~7,
  pp.~1069--1076, 2017.

\bibitem{PCAbio2}
J.~Lever, M.~Krzywinski, and N.~Altman, ``{Points of significance: Principal
  component analysis},'' {\em Nature methods}, vol.~14, no.~7, pp.~641--643,
  2017.

\bibitem{samadi2}
U.~Tantipongpipat, S.~Samadi, M.~Singh, J.~H. Morgenstern, and S.~Vempala,
  ``Multi-criteria dimensionality reduction with applications to fairness,''
  {\em Advances in neural information processing systems}, vol.~32, 2019.

\bibitem{pareto}
M.~M. Kamani, F.~Haddadpour, R.~Forsati, and M.~Mahdavi, ``Efficient fair
  principal component analysis,'' {\em Machine Learning}, pp.~1--32, 2022.

\bibitem{ami}
G.~Zalcberg and A.~Wiesel, ``Fair principal component analysis and filter
  design,'' {\em IEEE Transactions on Signal Processing}, vol.~69,
  pp.~4835--4842, 2021.

\bibitem{RPCA}
N.~Kwak, ``{Principal component analysis based on L1-norm maximization},'' {\em
  IEEE transactions on Pattern Analysis and Machine Intelligence}, vol.~30,
  no.~9, pp.~1672--1680, 2008.

\bibitem{sparsePCA1}
H.~Zou, T.~Hastie, and R.~Tibshirani, ``{Sparse principal component
  analysis},'' {\em {Journal of Computational and Graphical Statistics}},
  vol.~15, no.~2, pp.~265--286, 2006.

\bibitem{sparsePCA2}
H.~Zou and L.~Xue, ``{A Selective Overview of Sparse Principal Component
  Analysis},'' {\em Proceedings of the IEEE}, vol.~106, no.~8, pp.~1311--1320,
  2018.

\bibitem{sparsePCA3}
A.~d'Aspremont, L.~Ghaoui, M.~Jordan, and G.~Lanckriet, ``{A direct formulation
  for sparse PCA using semidefinite programming},'' {\em {Advances in Neural
  Information Processing Systems}}, vol.~17, 2004.

\bibitem{mohamad}
M.~M. Naghsh, M.~Masjedi, A.~Adibi, and P.~Stoica, ``Max--min fairness design
  for {MIMO} interference channels: A minorization--maximization approach,''
  {\em IEEE Transactions on Signal Processing}, vol.~67, no.~18,
  pp.~4707--4719, 2019.

\bibitem{pulmar}
L.~Zhao and D.~P. Palomar, ``Maximin joint optimization of transmitting code
  and receiving filter in radar and communications,'' {\em IEEE Transactions on
  Signal Processing}, vol.~65, no.~4, pp.~850--863, 2016.

\bibitem{MM}
Y.~Sun, P.~Babu, and D.~P. Palomar, ``Majorization-minimization algorithms in
  signal processing, communications, and machine learning,'' {\em IEEE
  Transactions on Signal Processing}, vol.~65, no.~3, pp.~794--816, 2016.

\bibitem{boyd}
S.~Boyd and L.~Vandenberghe, {\em Convex optimization}.
\newblock Cambridge university press, 2004.

\bibitem{cvx}
M.~Grant and S.~Boyd, ``{CVX}: Matlab software for disciplined convex
  programming, version 2.1,'' 2014.

\bibitem{sion}
M.~Sion, ``On general minimax theorems.,'' {\em Pacific Journal of
  Mathematics}, vol.~8, no.~1, pp.~171--176, 1958.

\bibitem{von}
A.~W. Marshall, I.~Olkin, and B.~C. Arnold, {\em Inequalities: theory of
  majorization and its applications}.
\newblock Springer, 1979.

\bibitem{recht}
B.~Recht, M.~Fazel, and P.~A. Parrilo, ``Guaranteed minimum-rank solutions of
  linear matrix equations via nuclear norm minimization,'' {\em SIAM review},
  vol.~52, no.~3, pp.~471--501, 2010.

\end{thebibliography}

\end{document}